\documentclass{article}

\usepackage{amsmath} 
\usepackage{amssymb}
\usepackage{graphicx}
\usepackage{siunitx}
\usepackage{booktabs}
\usepackage{float}
\usepackage{xcolor}
\usepackage{caption}
\usepackage{subcaption}
\usepackage{bm}
\usepackage{array}    % For advanced table features
\usepackage{placeins}
\usepackage{amsmath,amsfonts,bm}

\def\eqref#1{equation~\ref{#1}}
\def\1{\bm{1}}

\DeclareMathAlphabet{\mathsfit}{\encodingdefault}{\sfdefault}{m}{sl}
\SetMathAlphabet{\mathsfit}{bold}{\encodingdefault}{\sfdefault}{bx}{n}

\usepackage{algorithm}
\usepackage{algpseudocode} 
\usepackage[preprint]{neurips_2025}

\usepackage[utf8]{inputenc} % allow utf-8 input
\usepackage[T1]{fontenc}    % use 8-bit T1 fonts
\usepackage{hyperref}       % hyperlinks
\usepackage{url}            % simple URL typesetting
\usepackage{booktabs}       % professional-quality tables
\usepackage{amsfonts}       % blackboard math symbols
\usepackage{nicefrac}       % compact symbols for 1/2, etc.
\usepackage{microtype}      % microtypography
\usepackage{xcolor}         % colors

\usepackage{tikz}
\usepackage{graphicx}
\usetikzlibrary{shapes.geometric, arrows.meta, positioning,decorations.pathreplacing, positioning}
\usetikzlibrary{patterns}

\title{Video Editing for Audio-Visual Dubbing}

\author{%
  Binyamin Manela\\
  Faculty of Engineering\\
  Bar-Ilan University\\
  \texttt{manelab@biu.ac.il} \\
  \And 
  Sharon Gannot\\
  Faculty of Engineering\\
  Bar-Ilan University\\
  \texttt{sharon.gannot@biu.ac.il} \\
  \And 
  Ethan Fetyaya\\
  Faculty of Engineering\\
  Bar-Ilan University\\
  \texttt{ethan.fetaya@biu.ac.il} \\
}

\begin{document}

\maketitle

\begin{abstract}
Visual dubbing, the synchronization of facial movements with new speech, is crucial for making content accessible across different languages, enabling broader global reach. However, current methods face significant limitations. Existing approaches often generate talking faces, hindering seamless integration into original scenes, or employ inpainting techniques that discard vital visual information like partial occlusions and lighting variations. This work introduces EdiDub, a novel framework that reformulates visual dubbing as a content-aware editing task. EdiDub preserves the original video context by utilizing a specialized conditioning scheme to ensure faithful and accurate modifications rather than mere copying. On multiple benchmarks, including a challenging occluded-lip dataset, EdiDub significantly improves identity preservation and synchronization. Human evaluations further confirm its superiority, achieving higher synchronization and visual naturalness scores compared to the leading methods. These results demonstrate that our content-aware editing approach outperforms traditional generation or inpainting, particularly in maintaining complex visual elements while ensuring accurate lip synchronization.
\end{abstract}

% \begin{abstract}
%   Visual dubbing, i.e., synchronizing facial movements with new speech, remains a fundamental challenge in video synthesis, with applications spanning entertainment, education, and accessibility. Existing approaches either treat the problem as talking-face generation, struggling with contextual integration, or as pure inpainting that discards valuable visual information. Our goal is to stay as faithful to the original video as possible, while providing good lip synchronization. To achieve this goal, we first formulate the task as a video editing problem, instead of an inpainting problem, where our model has full access to the original video. This, however, can cause problems during training, as we need to prevent the model from merely reproducing the source video's lip movements rather than generating appropriate new ones. In order to avoid this pitfall, we designed a novel conditioning signal that maximizes visual cues accessibility while minimizing leakage of the original lip movement. To validate the usefulness of this approach, we curated a small test benchmark of challenging videos where the lips are partially occluded during speech. We then show on this benchmark, and on other standard benchmarks, that our approach provides better results compared to existing methods. 
% \end{abstract}

\begin{figure*}[t]
  \centering
  \includegraphics[width=\linewidth]{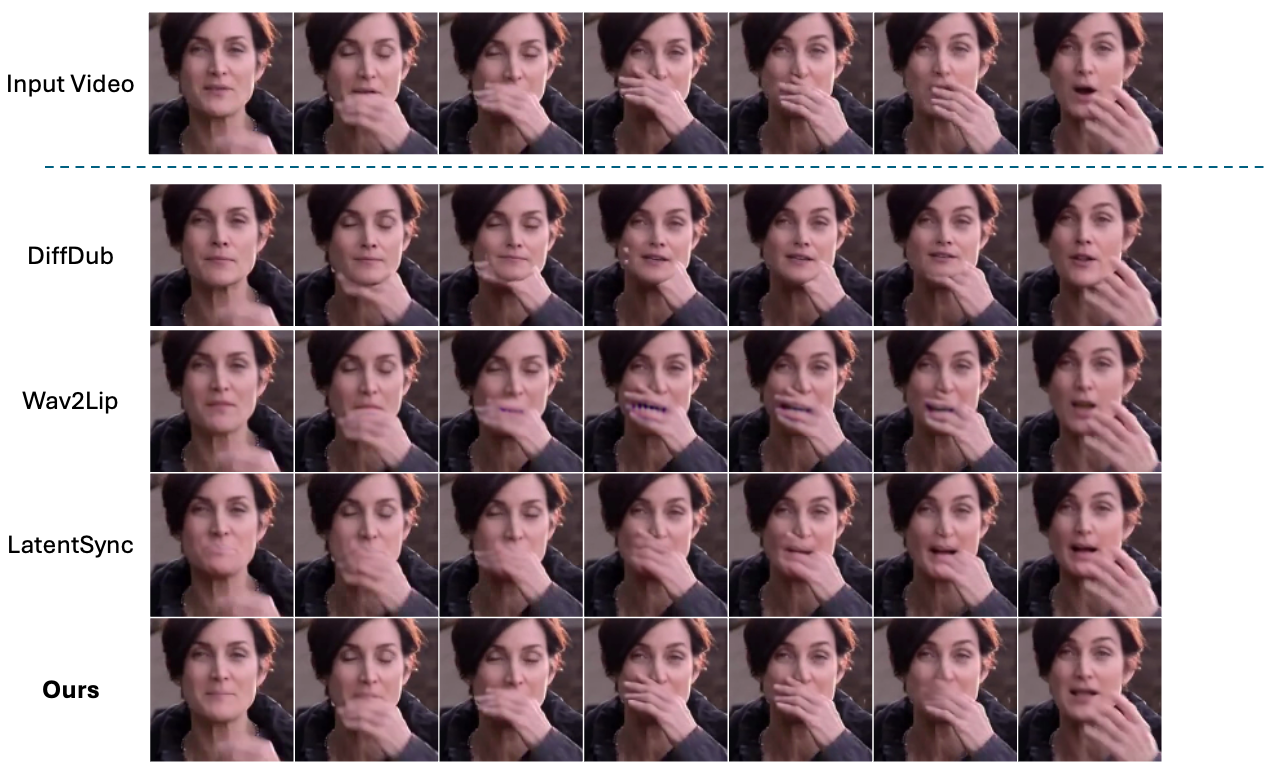}
  \caption{\textbf{Qualitative comparison across time.} Each row presents 7 consecutive frames from a different method, with the original video at the top and our model at the bottom. Existing methods create severe artifacts on these examples where the mouth is occluded.}
  \label{fig:qual_comparison}
\end{figure*}

\section{Introduction}
\label{sec:intro}
In an increasingly globalized media landscape, the seamless integration of speech and visual content across language barriers remains a formidable challenge. Every day, millions of viewers encounter dubbed content where visual lip movements fail to align with translated audio, creating a jarring disconnect that undermines immersion and comprehension.
This visual-auditory mismatch, often called the “McGurk effect” \citep{tiippana2014mcgurk}, is not merely an aesthetic concern but a fundamental barrier to effective cross-cultural communication and content accessibility.

Recent advancements in generative models for visual content have made visual dubbing an attainable and technically viable goal. This process of realistically synchronizing facial movements with new speech has spurred rapid development of effective, real-world tools. This has important applications, as Hollywood studios, for example, spend millions annually on dubbing international releases. Visual dubbing can also help democratize global content distribution, allowing independent creators and small-budget productions to make their content available across multiple languages. In each context, the believability of the synthesis directly impacts user engagement and information transfer.

Despite significant advances in deep generative models, particularly diffusion models \citep{ho2020denoising}, creating photorealistic dubbed videos with precise lip synchronization presents unique challenges that standard generative approaches struggle to address. The task demands not only visual fidelity but also precise temporal alignment, identity preservation, and contextual awareness - all while maintaining natural facial dynamics across diverse speaking styles and environments.

Existing approaches to this problem generally fall into one of two categories, each with significant limitations. The first treats lip synchronization as a talking-face generation problem \citep{tan2024flowvqtalker,stypulkowski2024diffused,zhang2024emotalker}, focusing on producing realistic and convincing videos but often failing to blend seamlessly within an existing visual scene. These methods typically generate impressive isolated results but are not suitable for blending within an existing context, as often required in real-world dubbing scenarios. The second category approaches the task as a pure inpainting problem, erasing the mouth region entirely and reconstructing it from scratch based on audio cues and reference images \citep{li2024latentsync,prajwal2020lip,liu2024diffdub}. While this simplifies the generation procedure, it discards valuable information present in the original footage, such as occlusions, motion blur, lighting variations, and shadow structures, that cannot be reliably inferred from reference frames and audio alone.

In this work, we present a novel diffusion-based dubbing framework, named \textbf{EdiDub}, that re-formulates the lip synchronization problem. Rather than approaching it as either generation or inpainting, we formulate it as a content-aware editing task aiming to preserve and leverage all available visual information. To maintain full access to the input clip during dubbing, we condition our model on the full original video at inference time. This, however, presents a problem during training, when the conditioning video is also the target, as this can lead to simply copying the original video. We therefore devised a new training process that uses a novel conditioning scheme, where for each frame we find a suitable reference frame that is most suitable for training. This allows us to pass on valuable spatial information while avoiding degenerate solutions and minimizing domain shift between training and inference. While our conditioning scheme contains all the necessary information, the generated output might still vary from the original video. We therefore also use inversion \citep{song2020denoising} to initialize the generation process. This further enhances faithfulness to the visual context, as we start our editing with a close recreation of the original video.

Complementary to the identity and visual-context preservation, we provide the model with a speech representation, using HuBERT features \citep{hsu2021hubert}, to ensure that the generated lip movements are synchronized with the input audio. By exploiting the known and constant temporal relation between the audio features and the video frames, we were able to simplify the model and avoid cross-attention layers as used by previous methods \citep{li2024latentsync}, while preserving the full expressiveness of the model. 

As is common practice in image and video generation directly on the pixel space, we reduce computational cost by first generating a low-resolution video, followed by a super-resolution phase to upscale it \citep{ho2022video, ho2022cascaded}. However, we found that existing video super-resolution (VSR) models performed poorly in our setting. To address this, we developed a dedicated super-resolution approach tailored to the specific characteristics of the dubbing task. Notably, only the modified lip region requires upscaling, and this prior knowledge allows us to simplify the super-resolution process while receiving high-fidelity and accurate results.

To evaluate our algorithm, we curated two benchmarks with videos selected from VoxCeleb2 \citep{chung2018voxceleb2} representing front-facing and occluded videos, to measure how these models handle the easy and hard cases. We also present two novel metrics to evaluate the naturalness and faithfulness of the dubbed videos by measuring the identity alignment with the original frames and identity consistency over time. Finally, we evaluate our model using Mean Opinion Score (MOS) from human annotators, evaluating naturalness and synchronization. 

Our comprehensive evaluations on multiple benchmark datasets, including a particularly difficult benchmark of videos with partial occlusions over the lip area, demonstrate that this editing approach significantly outperforms existing methods with high visual dubbing quality. We show, for example, in Fig.~\ref{fig:qual_comparison}, how our framework handles videos where the lips are partially occluded, while other methods create significant artifacts. 
Quantitative metrics show significant improvements in both visual fidelity and lip-sync accuracy compared to previous methods, while qualitative assessments confirm the system’s ability to handle challenging real-world scenarios, including partial occlusions and complex lighting conditions.

\section{Related Work}
\label{sec:related_work}
Generative models for visual dubbing evolved through two main paradigms: 

\vspace{3 pt}
\noindent\textbf{GAN-based approaches} pioneered the field, with Wav2Lip \citep{prajwal2020lip} using a pre-trained sync discriminator to guide mouth generation. Despite strong audio-visual correspondence, these methods struggle with smooth blending of generated area, produce blurry details at higher resolutions, and struggle with temporal consistency. StyleSync \citep{guan2023stylesync} improved quality through StyleGAN2 \citep{karras2020analyzing}, but limitations in blending quality and temporal consistency persisted. 

\vspace{3 pt}
\noindent\textbf{Diffusion-based methods} have demonstrated superior visual quality while tackling the challenge of temporal consistency. Diff2Lip \citep{mukhopadhyay2024diff2lip} applied frame-level diffusion to inpaint the lower face, conditioning on reference images and target audio while enforcing temporal coherence through sequence adversarial loss. DiffDub \citep{liu2024diffdub} addressed temporal issues with a two-stage approach: A diffusion autoencoder model for inpainting the lip area conditioned on frame embeddings, and a separate transformer for predicting the embedding sequence from reference images and HuBERT speech features. LatentSync \citep{li2024latentsync} leveraged latent diffusion models' \citep{rombach2022high} compact representation with supplementary pixel-level objectives (sync and fidelity losses), demonstrating that synchronization loss combined with a shared initial noise can reduce temporal inconsistencies despite not eliminating them completely. \newline

Critically, all these approaches formulate visual dubbing as either \textit{generation} or \textit{inpainting}, completely erasing the mouth region and reconstructing from scratch.
%This discards valuable contextual information such as occlusions, motion blur, and lighting variations that cannot be reliably inferred from reference frames alone.
In contrast, our {EdiDub} approach reformulates lip synchronization as a \textit{content-aware editing task}, aiming to preserve all available visual information. By conditioning on the full original video and using diffusion-based inversion to initialize generation, we maintain visual context fidelity while ensuring audio synchronization. This significantly enhances performance in challenging scenarios where inpainting methods struggle to recreate realistic details.

\section{Method}
\label{sec:method}

\begin{figure}[t]
  \centering

  \begin{subfigure}[b]{0.48\linewidth}
    \centering
    \fcolorbox{white}{gray!10}{
      \includegraphics[width=0.95\linewidth]{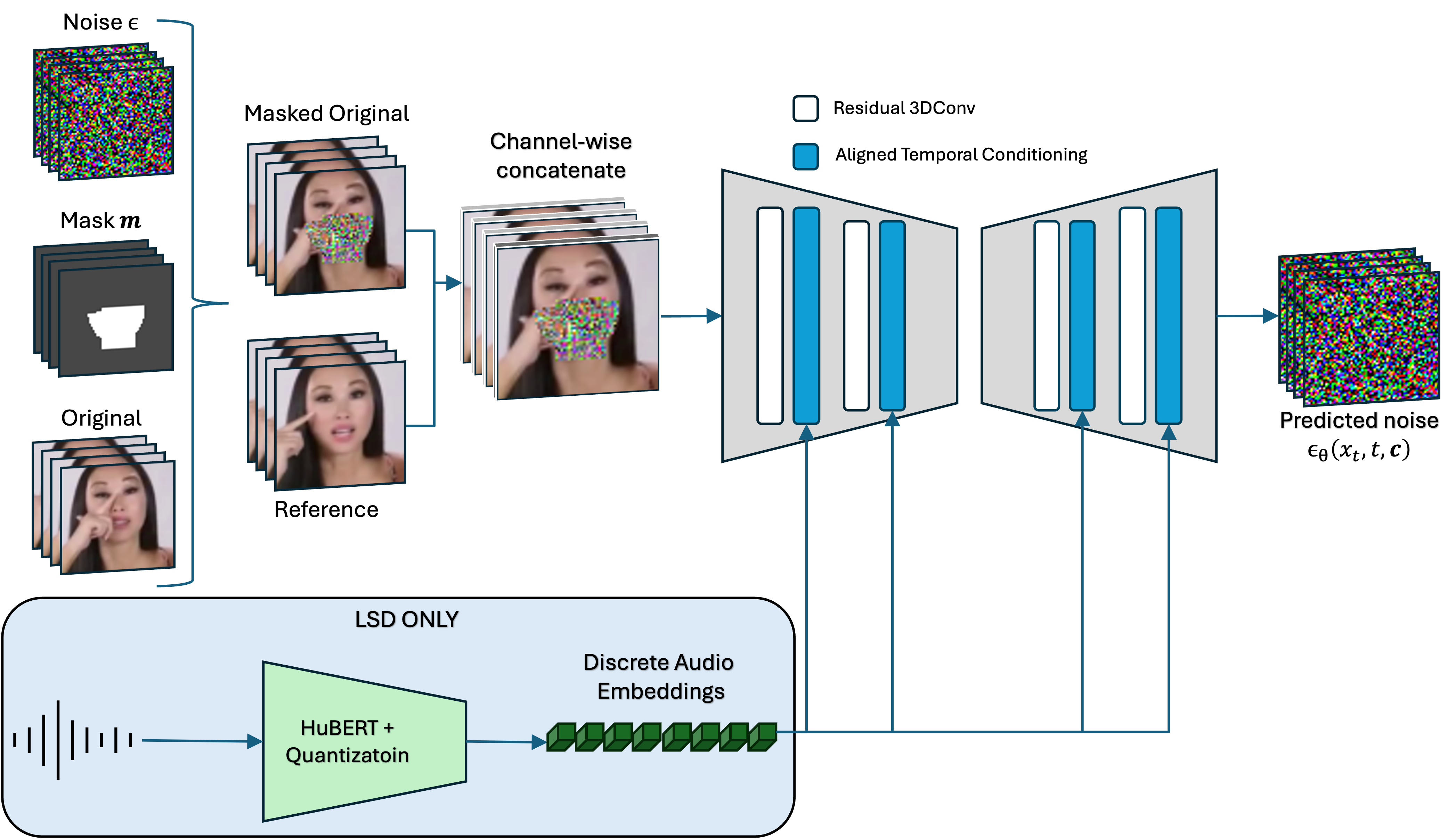}
    }
    \caption{Architectures of Lip-Sync Diffusion (LSD) and Super-Resolution Diffusion (SRD). LSD edits lip regions at $64{\times}64$; SRD restores details at $224{\times}224$.}
    \label{fig:pipeline_lsd}
  \end{subfigure}
  \hfill
  \begin{subfigure}[b]{0.48\linewidth}
    \centering
    \fcolorbox{white}{gray!10}{
      \includegraphics[width=0.95\linewidth]{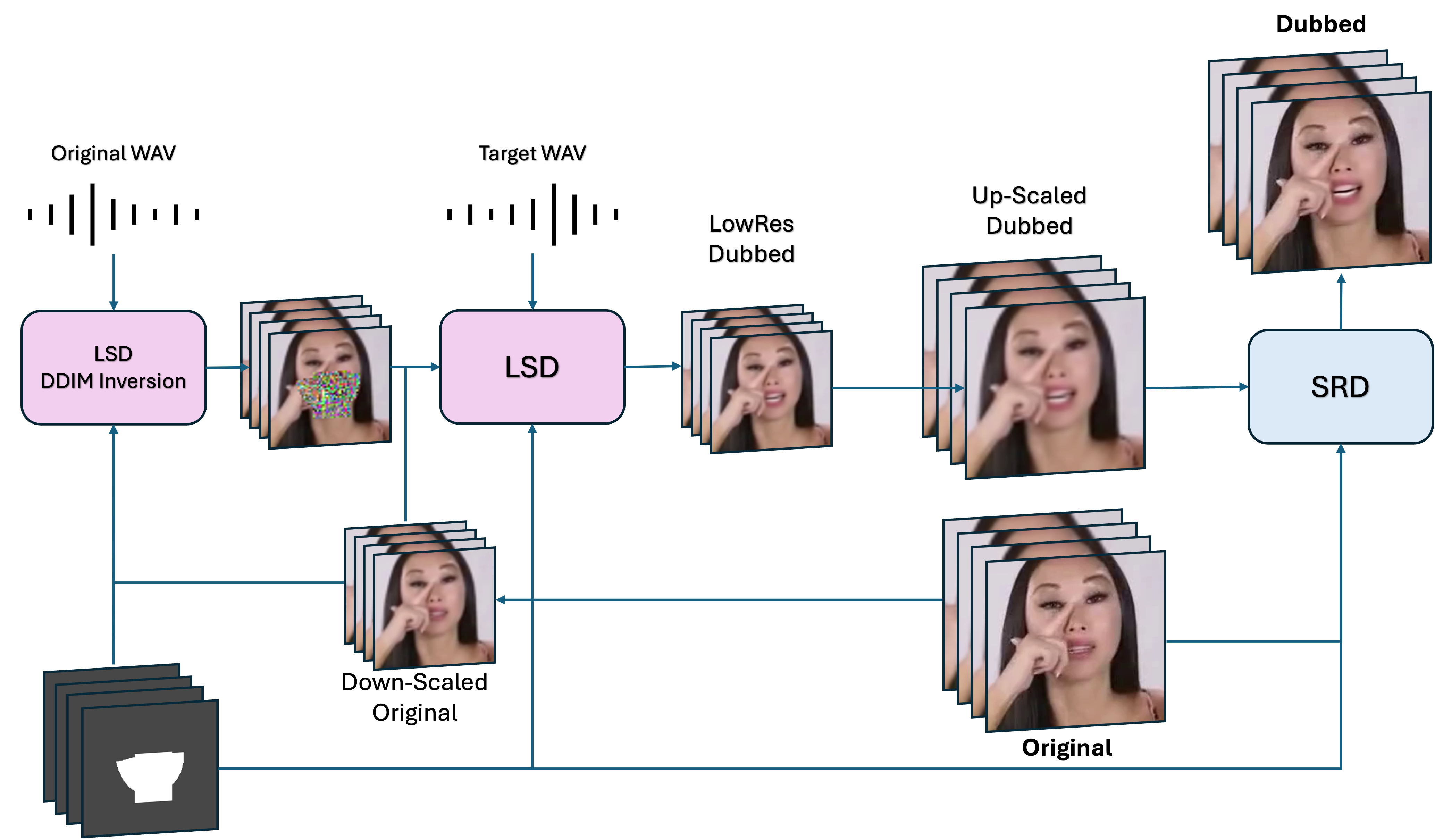}
    }
    \caption{Full inference pipeline: LSD generates aligned lip motion, followed by SRD producing a photorealistic output.}
    \label{fig:pipeline_full}
  \end{subfigure}

  \caption{\textbf{Overview of our dubbing system EdiDub.}  
  (a) Model architectures for both stages of our pipeline.  
  (b) End-to-end inference process for lip-synced facial video generation.}
  \label{fig:pipeline}
\end{figure}

In this section, we thoroughly describe our novel dubbing framework, named {EdiDub}. An overview of our method is illustrated in Fig.~\ref{fig:pipeline}. Our approach first performs \emph{content-aware editing} of a low-resolution clip to synchronize lip movements with a new speech track, then restores full spatial detail through a dedicated super-resolution diffusion model. Both stages are trained as \emph{masked video inpainting problems}.% and share a conditional 3D~U-Net backbone~\citep{cciccek20163d}, enabling efficient reuse of architectural components while dedicating each stage to a clearly delimited sub-task.

% ------------------------------------------------------------------
\subsection{Lip-Sync Diffusion (LSD)}
\label{subsec:lsd}

\paragraph{Masked DDPM objective.}
Let $\mathbf{x}_0,\mathbf{m}\!\in\!\mathbb{R}^{25\times64\times64\times3}$ be a 25-frame RGB clip and the binary mask for the facial area below the speaker's eyes, respectively. 
During the forward diffusion process, we add Gaussian noise \emph{only} inside the masked region~\citep{ho2020denoising}, producing $\mathbf{x}_t$ at timestep $t$. 
Given conditioning signals $\mathbf{c}$ (detailed below), the network predicts the added noise $\boldsymbol{\epsilon}_\theta$. Training the diffusion model as an inpainting model, rather than a regular generation model, provides faster and better convergence properties.
We normalize the reconstruction error by the number of masked pixels, ensuring that the loss remains comparable across faces of different sizes:
\begin{equation}
\mathcal{L}_{\mathrm{LSD}} \;=\;
  \mathbb{E}_{\mathbf{x}_0,\boldsymbol{\epsilon},t}
  \Bigg[
    \frac{\smash{\|
      (\boldsymbol{\epsilon}\!-\!\boldsymbol{\epsilon}_\theta(\mathbf{x}_t,t,\mathbf{c})) 
      \odot \mathbf{m}\|_2^2}}
         {\displaystyle \sum\nolimits_{i}\! m_i}
  \Bigg],
\label{eq:lsd_loss}
\end{equation}
where $\odot$ denotes element-wise multiplication and $m_i$ represents individual mask elements.

\paragraph{Conditioning signals.}
Our model incorporates a conditioning scheme tailored for the dubbing task.

% dubbing-oriented conditioning mechanisms that enable precise lip synchronization while maintaining visual fidelity and accurate reconstruction.

\paragraph{1. Accurate visual reference.} At inference time, we condition on the original unmasked video to capture valuable visual cues about the speaker's identity and appearance, including information about occlusions, lighting, and other temporally varying elements. However, na\"{i}vely using the original video during training creates a significant challenge. As the unmasked video serves as both the target output and the conditioning, it may result in a degenerate solution that learns to simply copy the input, resulting in poor results during inference, when audio and video are not synchronized.

To address this challenge, we developed an alternative reference frame selection strategy. For each frame in the sequence, we search for a same-identity \emph{reference frame} that minimizes the $\ell_2$ distance between facial landmarks, excluding the lips. To prevent trivial matches and avoid data leakage, we enforce a $\pm5$-frame exclusion window around the current frame. This approach retains useful visual information about the speaker while teaching the model to ignore the lip movements in the reference frames, thereby reducing the mismatch between training and inference. Concatenating this reference image with the input doubles the input channels to \mbox{$\mathbb{R}^{25\times64\times64\times6}$}, allowing the network to copy high-frequency spatial details instead of inferring them. Our experiments demonstrate that the reference selection for spatial proximity is crucial for the 3D convolutional layers to effectively deduce the facial structure, as shown in our ablation studies, presented in Sec. \ref{subsec:ablation}.

\paragraph{2. Quantized speech representation.} To represent the input speech we encode the audio using HuBERT~\citep{hsu2021hubert} and quantize the resulting features into $k\!=\!200$ discrete clusters obtained by running $k$-means on {one million} vectors drawn from LJ-Speech~\citep{ljspeech17}, a single-speaker corpus. Training the codebook on a single voice effectively nullifies the speaker-identity dimensions in the latent space. It removes speaker-specific directions from the latent space and isolates purely phonetic information that directly correlates with mouth motion, as shown by \citep{hsu2023revise}.
The resulting sequence $u_t\!\in\!\{1,\dots,200\}$, with a latents frame-rate of \SI{50}{\Hz}, is embedded via a learnable table $\mathbf{E}\!\in\!\mathbb{R}^{201\times d}$ (where the additional dimension accommodates a null symbol) and reshaped from $[T\!\times\!2,d]$ to $[T,2d]$ 
to align with the video frame-rate of \SI{25}{\Hz}. This representation enables precise phonetic-to-visual mapping independent of speaker identity or audio characteristics.

Furthermore, unlike in text-to-video models, where the relation between frames and tokens is inherently ambiguous, our setting benefits from an exact and known alignment between audio units and video frames. This allows us to bypass computationally expensive cross-attention layers in favor of a more efficient approach. We implement a depth-wise 1-D convolution followed by a linear projection that, given the condition $\mathbf{c}$, produces per-frame scale and shift vectors $(\gamma,\beta)$, which modulate the intermediate activations $\mathbf{h}$ via Adaptive Instance Normalization~\citep{huang2017arbitrary}:
\[
\operatorname{AdaIN}(\mathbf{h}) \;=\; \gamma(\mathbf{c})\!\cdot\!\mathbf{h}\;+\;\beta(\mathbf{c}).
\]
This lightweight alternative significantly reduces both parameter count and inference time while maintaining expressive capacity for precise temporal control.

\paragraph{DDIM inversion for targeted editing.}
While our conditioning scheme provides the model with all the needed information to reconstruct the masked region, we observed results that were not faithful enough to the original. To address this issue, we leverage DDIM-Inversion~\citep{song2020denoising} to {invert the observed clip} into its corresponding initial noise, effectively preserving its underlying structure:
 
\begin{equation}
\mathbf{z}_{t+1} =
\frac{1}{\sqrt{\alpha_{t+1}}}
\left(
  \mathbf{x}_t - \sqrt{1 - \alpha_{t+1}} \cdot \boldsymbol{\epsilon}_\theta(\mathbf{x}_t, t, \mathbf{c}_{\text{orig}})
\right),
\label{eq:ddim_inv}
\end{equation}
\noindent
starting from $\mathbf{x}_0$ and conditioned on the original speech, $\mathbf{c}_{\text{orig}}$. 
After $T\!=\!50$ inversion steps, we obtain the inverted latent code $\mathbf{z}_T^\star$, which serves as the initialization for the denoising process. During generation, we replace $\mathbf{c}_{\text{orig}}$ with $\mathbf{c}_{\text{new}}$, the tokens representation of the target speech, and apply classifier-free guidance~\citep{ho2022classifier} to enhance the fidelity of the lip synchronization:
\[
\tilde{\boldsymbol{\epsilon}}_\theta
  \;=\;(1+s)\,\boldsymbol{\epsilon}_\theta(\mathbf{x}_t,t,\mathbf{c}_{\text{new}})
     -s\,\boldsymbol{\epsilon}_\theta(\mathbf{x}_t,t,\varnothing),
\quad s\!=\!5.
\]
This procedure aims to {preserve everything except the lip motion}, maintaining fine details such as facial hair, motion blur, or occlusions that would otherwise be lost with traditional inpainting.

\paragraph{Long-video synthesis via MultiDiffusion.}
To efficiently and smoothly process arbitrarily long video sequences, we deploy MultiDiffusion~\citep{bar2023multidiffusion}. The video is split into overlapping temporal segments that are generated in parallel. After each diffusion step, we average the different predictions of the overlapping frames. 

% ------------------------------------------------------------------
\subsection{Super-Resolution Diffusion (SRD)}
\label{subsec:srd}

\paragraph{Input representation and inference.}
The SRD model extends our framework by restoring spatial detail from the low-resolution LSD output while preserving lip synchronization and temporal coherence. Following the same principles as LSD, it operates on temporal sequences using the MultiDiffusion strategy. % with identical window size (24 frames) and stride (12 frames) parameters, ensuring consistent overlap blending as described in~\eqref{eq:multidiffusion_avg}.
At each denoising step, SRD receives as input a concatenated tensor $\mathbf{y}_0 \in \mathbb{R}^{25\times224\times224\times6}$ comprising two components: (1) The masked high-resolution ground-truth frames $\mathbf{y}^{\text{GT}}_0 \in \mathbb{R}^{25\times224\times224\times3}$, and (2) A bicubic-upsampled low-resolution reference $\mathbf{y}^{\text{LR}}_0 \in \mathbb{R}^{25\times224\times224\times3}$

During training, $\mathbf{y}^{\text{LR}}_0$ is constructed by downsampling the ground truth frames and then upsampling them back to $224{\times}224$ resolution using bicubic interpolation. This provides a structurally accurate but detail-deficient version of the target, teaching the model to focus on enhancing facial details rather than modifying the overall structure. At inference time, we use the LSD output as the low-resolution source instead, upsampling it to match the target resolution and conditioning SRD to refine the generated output with high-frequency details.

\paragraph{Architecture adaptations.}
The SRD model builds upon the same conditional 3D U-Net backbone used in LSD, but introduces several key adaptations tailored for the super-resolution task. Naturally, it operates at a higher spatial resolution of $224{\times}224$ pixels. To accommodate memory constraints, the temporal context is reduced by training on 5-frame windows instead of 25. The network topology is also adjusted, featuring a deeper architecture with narrower channel widths. Additionally, the spatio-temporal self-attention layers present in LSD are omitted to simplify the model. These modifications strike a balance between computational efficiency and representational power. Notably, the fully convolutional design allows SRD to generalize to varying inference windows without necessitating retraining or architectural changes.

\paragraph{Training objective.}
We train the SRD model with the same masked‐reconstruction loss used in our low‐resolution lip‐sync diffusion (LSD) model (\eqref{eq:lsd_loss}), but computed at full spatial resolution. During training, we introduce both spatial and temporal variety by (1) randomly sampling the low‐resolution input size to be between 32 and 64 pixels, and (2) with $5\%$ probability replacing a conditioning frame with its predecessor. By sharing the reconstruction objective and employing these data augmentations, the two stages learns complementary behaviors that combine into a single, end‐to‐end framework for high‐fidelity video dubbing.

\section{Experiments}
\label{sec:experiments}
In this section, we evaluate our dubbing framework for visual dubbing against existing methods across multiple benchmarks. We first describe our experimental setup, introducing the datasets, baselines, and evaluation metrics. We then present quantitative results demonstrating the effectiveness of our approach, followed by comprehensive ablation studies that validate our design choices. For reproducibility, the code will be made available in the supplementary material.

\subsection{Experimental Setup}
\label{subsec:setup}

\paragraph{Datasets.} We evaluate our method on four benchmarks that span a range of difficulty levels. The first is \textbf{LRS2} \citep{son2017lip}, a large-scale English audio-visual dataset containing over 45,000 spoken sentences from BBC broadcasts, featuring substantial variation in head poses and speaking styles. The second is \textbf{LRS3} \citep{afouras2018lrs3}, an extension of LRS2 with more diverse content from TED and TEDx talks, providing challenging examples with expressive speaking patterns. We also use two new benchmarks curated from  VoxCeleb2 \citep{chung2018voxceleb2}. The first is  \textbf{Vox2-Front}, where facial orientations remain within $\pm25$ degrees from the frontal position, representing relatively controlled conditions within a real-world scenario. Finally, we include \textbf{Vox2-Occluded}, a challenging benchmark that contains videos with partial occlusions of the lip region that we have specifically designed to test the preservation of these occlusions during lip synchronization (see example in Fig. \ref{fig:qual_comparison}). The full pipeline used to generate this benchmark is provided in the Appendix. These datasets collectively provide a comprehensive testbed for evaluating visual dubbing methods across varying degrees of difficulty.

\paragraph{Baselines.} We compare our approach against three strong methods: Wav2Lip \citep{prajwal2020lip}, LatentSync \citep{li2024latentsync} and DiffDub \citep{liu2024diffdub}.
To ensure fair comparison, we used the official implementations and pre-trained models for all baselines. Our model was trained on LRS3, DiffDub was trained on HDTF \citep{zhang2021flow},  LatentSync was trained on both LRS3 and VoxCeleb, while Wav2Lip was trained on LRS2.

\paragraph{Evaluation Metrics.} We employ both quantitative metrics and human evaluations to assess the quality of generated videos: \textit{1) Lip Synchronization Accuracy:}
To evaluate synchronization accuracy, we utilize the state-of-the-art audio-video synchronization network, ModEFormer \citep{gupta2023modeformer}.
Motivated by the architecture of CLIP \citep{radford2021learning} and syncnet \citep{Chung16a}, ModEFormer embeds both audio and video into a shared, normalized, latent space, where the cosine similarity between the embeddings correlates to sync quality.
We use \textbf{LSE-D} (Lip Sync Error - Distance), which measures the average cosine distance between generated lips and target audio, and \textbf{LSE-C} (Lip Sync Error - Confidence), which evaluates the confidence score of the synchronization quality. Both of these metrics were suggested by \citep{prajwal2020lip}.

\textit{2) Visual Naturalness and Realism:}
While quantitative synchronization evaluation is well established, measuring the perceived naturalness and realism of a visually dubbed video is a challenging problem. Therefore, we designed two novel metrics tailored to measure dubbing quality utilizing a pretrained facial identity embedding network \citep{timesler}.
\textbf{ID-P} (Identity Preservation) calculates a frame-wise cosine distance between facial embeddings extracted from the original and generated videos. The motivation for this metric is that visual artifacts in the frame should cause a large difference between the embeddings. \textbf{ID-TC} (Identity Temporal Consistency) measures the cosine distance between embeddings of consecutive frames in the generated video. Lower values indicate smoother transitions and fewer temporal artifacts such as jitter or flicker. 

As a few poorly generated frames have a disproportionate impact on perceived quality, compared to their relative weight, we use a weighted average to calculate these metrics. Specifically, the averaging weights are computed using a softmax over the raw scores, allowing the metric to be more highly influenced by frames with visual artifacts. 

% Visual artifacts tend to have a disproportionate impact on perceived quality, compared to their frequency. To better align the metric with human perception, we introduce a simple attention mechanism that emphasizes bad frames. Specifically, the aggregation weights are computed using a softmax over the raw scores, allowing the metric to focus on visual artifacts and inaccuracies.

\textit{3) Human Evaluation:} For human evaluation, we employ two Mean Opinion Score (MOS) metrics: \textbf{MOS-Sync} (Synchronization), which collects human ratings on a scale of 1-10 assessing how well the lip movements match the audio, and \textbf{MOS-Nat} (Naturalness), which gathers human ratings on a scale of 1-10 evaluating the overall visual quality, absence of artifacts, and natural appearance of the generated videos. Full details in the appendix.

\subsection{Implementation Details}

Our LSD model operates on 25-frame sequences at 64×64 resolution, while the SRD model upsamples the resulting low-resolution video to 224×224 spatial resolution. Both models share a similar 3D U-Net architecture but with different capacities as described in Section \ref{subsec:srd}. We train both stages for 500,000 iterations using the AdamW optimizer \citep{loshchilov2017decoupled} with a 10,000 steps warmup, up to a learning rate of 1e-4, followed by a cosine-decay learning rate down to 1e-5, and a batch size of 28. Full implementation details are available in Appendix.

For audio processing, we extract HuBERT features at a rate of 50Hz (two units per video frame) and quantize them using our single-speaker codebook as detailed in Section \ref{subsec:lsd}. During inference, we use DDIM-inversion and DDIM-sampling with 50 steps and a classifier-free guidance scale of 5, on samples of 24 frames. For videos longer than 24 frames, we apply MultiDiffusion \citep{bar2023multidiffusion} with 50\% overlap between adjacent windows. For videos longer than 120 frames, we split the video into sections of 120 frames, with a 12-frame overlap. Each section is generated sequentially, with the leading 12 frames from the end of the previous section with no mask (mask of zeros for these frames), thus forcing the model to smoothly continue the last section. 
Inference was conducted on 1 NVIDIA A100 GPU (40 GB), and training was conducted on 1 NVIDIA H200 GPU. Training required approximately 5 days, while inference runs at approximately 2 frames per second.

\subsection{Quantitative Results}
\label{subsec:quant_results}

\begin{table}[t]
\centering
\caption{\textbf{Quantitative comparison} of our method against baselines across multiple datasets. Lower values are better for LSE-D, ID-P, and ID-TC, while higher values are better for LSE-C.}
\label{tab:quant_results}
\resizebox{\linewidth}{!}{%
\begin{tabular}{lccccccccccccccccc}
\toprule
& \multicolumn{4}{c}{\textbf{LRS2}} & \multicolumn{4}{c}{\textbf{LRS3}} & \multicolumn{4}{c}{\textbf{Vox2-Front}} & \multicolumn{4}{c}{\textbf{Vox2-Occluded}} \\
\cmidrule(lr){2-5} \cmidrule(lr){6-9} \cmidrule(lr){10-13} \cmidrule(lr){14-17}
Method & LSE-D↓ & LSE-C↑ & ID-P↓ & ID-TC↓ & LSE-D↓ & LSE-C↑ & ID-P↓ & ID-TC↓ & LSE-D↓ & LSE-C↑ & ID-P↓ & ID-TC↓ & LSE-D↓ & LSE-C↑ & ID-P↓ & ID-TC↓ \\
\midrule
Wav2Lip\citep{prajwal2020lip} & 0.734 & 0.021 & 0.049 & 0.045 & 0.666 & 0.048 & \textbf{0.045} & 0.048 & 0.651 & 0.039 & 0.046 & 0.027 & 0.666 & 0.028 & 0.048 & 0.040 \\
DiffDub\citep{liu2024diffdub} & 0.714 & 0.038 & 0.341 & 0.114 & 0.642 & 0.050 & 0.325 & 0.063 & 0.573 & 0.060 & 0.241 & 0.025 & 0.628 & 0.040 & 0.294 & 0.048 \\
LatentSync\citep{li2024latentsync} & 0.653 & 0.099 & 0.055 & 0.054 & 0.534 & 0.171 & 0.066 & 0.045 & 0.514 & 0.145 & 0.040 & \textbf{0.022} & 0.546 & 0.133 & 0.052 & 0.036 \\
Ours & \textbf{0.620} & \textbf{0.130} & \textbf{0.046} & \textbf{0.041} & \textbf{0.489} & \textbf{0.201} & 0.065 & \textbf{0.042} & \textbf{0.495} & \textbf{0.171} & \textbf{0.036} & \textbf{0.022} & \textbf{0.523} & \textbf{0.159} & \textbf{0.033} & \textbf{0.035} \\
\midrule
Real Videos & 0.591 & 0.109 & 0.000 & 0.034 & 0.497 & 0.167 & 0.000 & 0.043 & 0.513 & 0.130 & 0.000 & 0.020 & 0.558 & 0.103 & 0.000 & 0.034 \\
\bottomrule
\end{tabular}%
}
\end{table}

As shown in Table \ref{tab:quant_results}, our method consistently outperforms existing approaches across all datasets. For lip synchronization metrics, our approach achieves the lowest LSE-D scores and the highest LSE-C scores, indicating superior lip synchronization quality.
For identity preservation, our approach maintains competitive or superior performance with ID-P values and ID-TC values that are closest to those of real videos among all methods. This demonstrates our method's ability to preserve speaker identity while achieving accurate lip synchronization. Particularly on the Vox2-Occluded dataset, our approach shows significant improvements in identity preservation, with a $30.1\%$ reduction in ID-P compared to the second-best score, while maintaining high synchronization quality.
It is worth noting that our method's synchronization performances show competitive results, compared to real videos, suggesting that our synthesized videos achieve high-quality lip movements that are perceptually comparable to real recordings.  Standard errors of the results in Table \ref{tab:quant_results} are in the Appendix, as well as a link to video samples. 

\subsection{Qualitative Results and Human Evaluation}
\label{subsec:qual_results}
The human evaluation results in Table \ref{tab:human_eval} further validate the superior performance of our approach. Our method receives the highest MOS-Sync and MOS-Sync scores across all datasets, with scores on LRS3 that are statistically equivalent to  LatentSync.

The qualitative examples in Figure \ref{fig:qual_comparison} illustrate our method's ability to generate realistic lip movements while preserving speaker identity and handling challenging visual elements. 
Overall, both quantitative metrics and human evaluations consistently demonstrate that our approach outperforms existing methods in terms of lip synchronization quality, identity preservation, and visual naturalness across various datasets and challenging conditions.

\begin{table}[t]
\centering
\caption{\textbf{Human evaluation results} showing Mean Opinion Scores for synchronization quality (MOS-Sync) and naturalness (MOS-Nat) on a scale of 1-10. Higher is better. The table provides the average scores $\pm 1$ standard error} %Best results are highlighted in \textbf{bold}.}
\label{tab:human_eval}
\resizebox{\linewidth}{!}{
\begin{tabular}{lcccccccc}
\toprule
& \multicolumn{2}{c}{\textbf{LRS2}} & \multicolumn{2}{c}{\textbf{LRS3}} & \multicolumn{2}{c}{\textbf{Vox2-Front}} & \multicolumn{2}{c}{\textbf{Vox2-Occluded}} \\
\cmidrule(lr){2-3} \cmidrule(lr){4-5} \cmidrule(lr){6-7} \cmidrule(lr){8-9}
Method & MOS-Sync↑ & MOS-Nat↑ & MOS-Sync↑ & MOS-Nat↑ & MOS-Sync↑ & MOS-Nat↑ & MOS-Sync↑ & MOS-Nat↑ \\
\midrule
DiffDub\citep{liu2024diffdub} & 3.16 $\pm$ 0.44 & 4.81 $\pm$ 0.58 & 4.79 $\pm$ 0.47 & 7.19 $\pm$ 0.43 & 7.92 $\pm$ 0.42 & 8.41 $\pm$ 0.41 & 5.45 $\pm$ 0.49 & 7.27 $\pm$ 0.46 \\
Wav2Lip\citep{prajwal2020lip} & 5.07 $\pm$ 0.56 & 6.12 $\pm$ 0.53 & 6.75 $\pm$ 0.51 & 7.6 $\pm$ 0.47 & 7.07 $\pm$ 0.44 & 7.55 $\pm$ 0.43 & 6.92 $\pm$ 0.46 & 7.74 $\pm$ 0.44 \\
LatentSync\citep{li2024latentsync} & 5.71 $\pm$ 0.51 & 7.18 $\pm$ 0.5 & \textbf{7.59 $\pm$ 0.37} & \textbf{8.46 $\pm$ 0.36} & 7.93 $\pm$ 0.39 & 8.44 $\pm$ 0.37 & \textbf{8.37 $\pm$ 0.36} & 9.14 $\pm$ 0.27 \\
Ours & \textbf{7.56 $\pm$ 0.32} & \textbf{8.31 $\pm$ 0.27} & \textbf{7.54 $\pm$ 0.36} & \textbf{8.41 $\pm$ 0.29} & \textbf{8.98 $\pm$ 0.21} & \textbf{9.21 $\pm$ 0.22} & \textbf{8.50 $\pm$ 0.33} & \textbf{9.42 $\pm$ 0.16} \\
\midrule
Real Videos & 8.97 $\pm$ 0.26 & 9.35 $\pm$ 0.21 & 9.51 $\pm$ 0.27 & 9.69 $\pm$ 0.26 & 9.1 $\pm$ 0.35 & 9.45 $\pm$ 0.26 & 8.61 $\pm$ 0.32 & 8.89 $\pm$ 0.29 \\
\bottomrule
\end{tabular}
}
\end{table}

\subsection{Ablation Studies}
\label{subsec:ablation}

To validate our design choices, we conducted comprehensive ablation studies by systematically removing key components of our framework. Results are presented in Table \ref{tab:ablation}. % presents results on Vox2-Front and Vox2-Occluded.

\begin{table}[t]
\centering
\caption{\textbf{Ablation study results} on the Vox2-Front and Vox2-Occluded datasets, examining the impact of each key component in our framework. Lower values are better for LSE-D, ID-P, and ID-TC, while higher values are better for LSE-C.}
\label{tab:ablation}
\resizebox{\linewidth}{!}{%
\begin{tabular}{lcccccccc}
\toprule
& \multicolumn{4}{c}{\textbf{Vox2-Front}} & \multicolumn{4}{c}{\textbf{Vox2-Occluded}} \\
\cmidrule(lr){2-5} \cmidrule(lr){6-9}
Method & LSE-D↓ & LSE-C↑ & ID-P↓ & ID-TC↓ & LSE-D↓ & LSE-C↑ & ID-P↓ & ID-TC↓ \\
\midrule
Full model (ours) & 0.495 & 0.171 & \textbf{0.036} & \textbf{0.022} & 0.523 & 0.159 & \textbf{0.033} & \textbf{0.035} \\
\midrule
w/o DDIM inversion & 0.475 & 0.188 & 0.051 & 0.023 & \textbf{0.501} & 0.178 & 0.054 & 0.041 \\
w/o optimized reference & \textbf{0.468} & \textbf{0.195} & 0.065 & 0.025 & \textbf{0.501} & 0.183 & 0.089 & 0.042 \\
w/o inversion \& opt. reference & \textbf{0.469} & \textbf{0.195} & 0.124 & 0.026 & 0.499 & \textbf{0.185} & 0.200 & 0.039 \\
w/o HuBERT clustering & 0.544 & 0.167 & 0.041 & \textbf{0.022} & 0.578 & 0.138 & 0.37 & 0.036 \\
\bottomrule
\end{tabular}%
}
\end{table}

\paragraph{Impact of DDIM inversion.} Removing DDIM inversion considerably degrades identity preservation metrics, with ID-P increasing by $40\%$ on Vox2-Front and by $64\%$ on Vox2-Occluded. Interestingly, synchronization metrics (LSE-D and LSE-C) actually improve without DDIM inversion, which aligns with our hypothesis that there is a trade-off between identity and context preservation and lip synchronization. Without proper identity anchoring, the model relies more heavily on audio conditioning, leading to better lip sync at the cost of identity preservation.

\paragraph{Optimized reference selection.} When using random reference frames instead of our optimized selection strategy, identity preservation suffers further, with ID-P increases by $80\%$ on Vox2-Front and $170\%$ on Vox2-Occluded. This demonstrates that selecting appropriate reference frames is crucial for maintaining subject identity throughout generation. Similar to removing DDIM inversion, synchronization metrics continue to show slight improvements, reinforcing the trade-off between identity preservation and synchronization.

\paragraph{Combined effect of inversion and reference selection.} Removing both DDIM inversion and optimized reference selection leads to a dramatic degradation in identity preservation, with ID-P increasing substantially by a factor of $2.5$ on Vox2-Front and a factor of $6$ on Vox2-Occluded. This substantial drop confirms the synergistic relationship between these two mechanisms in preserving identity, with each addressing complementary aspects of the visual dubbing challenge, i.e., the reference frames provide the necessary visual information, while DDIM-Inversion helps utilize it effectively. The synchronization metrics remain similar to removing either component individually, suggesting a limit in synchronization numerical evaluation, using a dedicated expert network.

\paragraph{HuBERT clustering.} Removing clustering and conditioning our model on the continuous audio features significantly impacts synchronization metrics, with LSE-D increases by about $10\%$ on both Vox2-Front and Vox2-Occluded. LSE-C also degrades by $15\%$ on Vox2-Occluded. Identity preservation metrics show minimal changes compared to the full model, confirming that HuBERT clustering primarily affects the audio-visual synchronization aspects rather than identity preservation.

These ablation results clearly demonstrate that each component of our framework contributes meaningfully to the overall performance. The DDIM inversion and optimized reference selection predominantly impact identity preservation, while HuBERT clustering primarily influences synchronization.

\subsection{Limitations}
\label{subsec:limitations}
One limitation of our proposed work is the time required for dubbing. Our approach generates ~2 frames per second on a powerful GPU. While this is fine for many applications, further work should be made to improve computational efficiency. Furthermore, while we observed improvement over competing methods, there may still be some visual artifacts remaining. We conjecture that this is due to the fact that we trained a video diffusion model on a relatively modest-sized dataset and did not use a model pretrained on an enormous corpus. This was done to align with competing works that also trained their diffusion model on similar-sized datasets.

Another important consideration is the potential negative societal impact. While the goal is to allow shared content across languages, this technology can also be misused for spreading misinformation. However, we do note that generative models that create ``talking heads'' are already common, and the benefit of this specific approach (more faithful to the original exact motion) is more relevant in legitimate dubbing. 
% Despite the advancements presented, our method has several limitations worth noting. 
% The current implementation operates only at 25 FPS with fixed resolution, requiring preprocessing for videos with different specifications. 
% Our reliance on third-party facial landmark detection may propagate inaccuracies into the final output. 
% Temporal stability issues can arise during rapid facial movements or inconsistent face detection, particularly affecting the performance of 3D convolutional layers. 
% While our approach handles partial occlusions better than previous methods, extreme occlusions can still generate artifacts. 
% Performance may degrade under challenging lighting conditions or with noisy audio due to training data limitations. 
% Finally, for safety reasons, the model was neither trained nor tested on videos of minors, necessitating careful evaluation before application to such demographic groups.

\section{Conclusions}
\label{sec:conclusions}
In this work, we present a new approach for visual dubbing, based on video editing. Our main design philosophy is to remain as faithful as possible to the original video while synchronizing the lips with the target speech. This is especially important when the area around the lips contains temporal visual information that is not a result of the speech, nor appears similarly in other frames, for example, occlusions. Empirically, we show the usefulness of this approach on a variety of benchmarks, using quantitative metrics as well as human evaluation.  

\bibliography{dubbing_arxiv}
\bibliographystyle{unsrt}

%%%%%%%%%%%%%%%%%%%%%%%%%%%%%%%%%%%%%%%%%%%%%%%%%%%%%%%%%%%%
\newpage
\appendix

\section{Extended Results}
In this section, we provide a deeper look into the results of our experiments, including statistical tests, our MOS web page, and our MOS results analysis process. All the videos from the evaluations can be found on our anonymous GitHub website: \url{https://edidub.github.io/edidub-results/}.

\subsection{Quantitative Results}
\label{subsec:quant_results}
This section presents comprehensive quantitative comparisons across all evaluated benchmarks and algorithms. To assess statistical significance, we perform a one-sided Wilcoxon Signed-Rank Test \citep{woolson2005wilcoxon} comparing each competing method against our proposed algorithm. The null hypothesis assumes no performance difference between methods, while the alternative hypothesis posits that our algorithm achieves superior scores. Since all methods are evaluated on identical video sets, this paired nonparametric test effectively determines whether our approach significantly outperforms each baseline.

Statistical significance levels are indicated as follows: "ns" represents non-significant results ($p > 0.05$), "*" indicates $0.01<p\le0.05$, "**" denotes $0.001<p\le0.01$, and "***" signifies $p\le0.001$. All experimental results are reported as mean $\pm$ 1 standard error.

\FloatBarrier
\begin{table}[H]
\centering
\label{tab:quant_results_ext_LRS2}
\resizebox{\linewidth}{!}{%
\begin{tabular}{lccccc}
\toprule
& \multicolumn{4}{c}{\textbf{LRS2}} \\
\cmidrule(lr){2-5}
Method & LSE-D↓ & LSE-C↑ & ID-P↓ & ID-TC↓ \\
\midrule
Wav2Lip\citep{prajwal2020lip} & $0.734\pm0.011^{***}$ & $0.021\pm0.002^{***}$ & $0.049\pm0.005^{*}$ & $0.045\pm0.008^{***}$ \\
DiffDub\citep{liu2024diffdub} & $0.714\pm0.015^{***}$ & $0.038\pm0.006^{***}$ & $0.341\pm0.026^{***}$ & $0.114\pm0.021^{***}$ \\
LatentSync\citep{li2024latentsync} & $0.653\pm0.015^{***}$ & $0.099\pm0.006^{***}$ & $0.055\pm0.006^{**}$ & $0.054\pm0.012^{**}$ \\
Ours & $\mathbf{0.620\pm0.015}$ & $\mathbf{0.130\pm0.007}$ & $\mathbf{0.046\pm0.008}$ & $\mathbf{0.041\pm0.007}$ \\
\midrule
Real Videos & $0.591\pm0.015$ & $0.109\pm0.008$ & $0.000\pm0.000$ & $0.034\pm0.005$ \\
\bottomrule
\end{tabular}%
}
\end{table}

\FloatBarrier
\begin{table}[H]
\centering
\label{tab:quant_results_ext_LRS3}
\resizebox{\linewidth}{!}{%
\begin{tabular}{lccccc}
\toprule
& \multicolumn{4}{c}{\textbf{LRS3}} \\
\cmidrule(lr){2-5}
Method & LSE-D↓ & LSE-C↑ & ID-P↓ & ID-TC↓ \\
\midrule
Wav2Lip\citep{prajwal2020lip} & $0.666\pm0.006^{***}$ & $0.048\pm0.003^{***}$ & $0.045\pm0.002^{***}$ & $0.048\pm0.005^{***}$ \\
DiffDub\citep{liu2024diffdub} & $0.642\pm0.013^{***}$ & $0.050\pm0.008^{***}$ & $0.325\pm0.017^{***}$ & $0.063\pm0.008^{***}$ \\
LatentSync\citep{li2024latentsync} & $0.534\pm0.007^{***}$ & $0.171\pm0.005^{***}$ & $\mathbf{0.066\pm0.004}^{\mathrm{ns}}$ & $0.045\pm0.005^{***}$ \\
Ours & $\mathbf{0.489\pm0.007}$ & $\mathbf{0.201\pm0.004}$ & $\mathbf{0.065\pm0.006}$ & $\mathbf{0.042\pm0.004}$ \\
\midrule
Real Videos & $0.497\pm0.010$ & $0.167\pm0.008$ & $0.000\pm0.000$ & $0.043\pm0.005$ \\
\bottomrule
\end{tabular}%
}
\end{table}

\FloatBarrier
\begin{table}[H]
\centering
\label{tab:quant_results_ext_VoxFront}
\resizebox{\linewidth}{!}{%
\begin{tabular}{lccccc}
\toprule
& \multicolumn{4}{c}{\textbf{Vox2-Front}} \\
\cmidrule(lr){2-5}
Method & LSE-D↓ & LSE-C↑ & ID-P↓ & ID-TC↓ \\
\midrule
Wav2Lip\citep{prajwal2020lip} & $0.651\pm0.008^{***}$ & $0.039\pm0.003^{***}$ & $0.046\pm0.004^{***}$ & $0.027\pm0.002^{***}$ \\
DiffDub\citep{liu2024diffdub} & $0.573\pm0.011^{***}$ & $0.060\pm0.008^{***}$ & $0.241\pm0.014^{***}$ & $0.025\pm0.002^{**}$ \\
LatentSync\citep{li2024latentsync} & $0.514\pm0.008^{***}$ & $0.145\pm0.005^{***}$ & $0.040\pm0.004^{**}$ & $\mathbf{0.022\pm0.002}^{\mathrm{ns}}$ \\
Ours & $\mathbf{0.495\pm0.010}$ & $\mathbf{0.171\pm0.006}$ & $\mathbf{0.036\pm0.005}$ & $\mathbf{0.022\pm0.002}$ \\
\midrule
Real Videos & $0.513\pm0.009$ & $0.130\pm0.006$ & $0.000\pm0.000$ & $0.020\pm0.002$ \\
\bottomrule
\end{tabular}%
}
\end{table}

\FloatBarrier
\begin{table}[H]
\centering
\label{tab:quant_results_ext_VoxOccluded}
\resizebox{\linewidth}{!}{%
\begin{tabular}{lccccc}
\toprule
& \multicolumn{4}{c}{\textbf{Vox2-Occluded}} \\
\cmidrule(lr){2-5}
Method & LSE-D↓ & LSE-C↑ & ID-P↓ & ID-TC↓ \\
\midrule
Wav2Lip\citep{prajwal2020lip} & $0.666\pm0.008^{***}$ & $0.028\pm0.002^{***}$ & $0.048\pm0.005^{***}$ & $0.040\pm0.004^{***}$ \\
DiffDub\citep{liu2024diffdub} & $0.628\pm0.013^{***}$ & $0.040\pm0.008^{***}$ & $0.294\pm0.015^{***}$ & $0.048\pm0.006^{***}$ \\
LatentSync\citep{li2024latentsync} & $0.546\pm0.009^{***}$ & $0.133\pm0.004^{***}$ & $0.052\pm0.005^{***}$ & $\mathbf{0.036\pm0.004}^{\mathrm{ns}}$ \\
Ours & $\mathbf{0.523\pm0.010}$ & $\mathbf{0.159\pm0.005}$ & $\mathbf{0.033\pm0.004}$ & $\mathbf{0.035\pm0.004}$ \\
\midrule
Real Videos & $0.558\pm0.010$ & $0.103\pm0.005$ & $0.000\pm0.000$ & $0.034\pm0.004$ \\
\bottomrule
\end{tabular}%
}
\end{table}

\subsection{Mean Opinion Score (MOS) Evaluation}
\begin{figure}[t]
\centering
\begin{subfigure}[b]{0.8\linewidth}
\includegraphics[width=\linewidth]{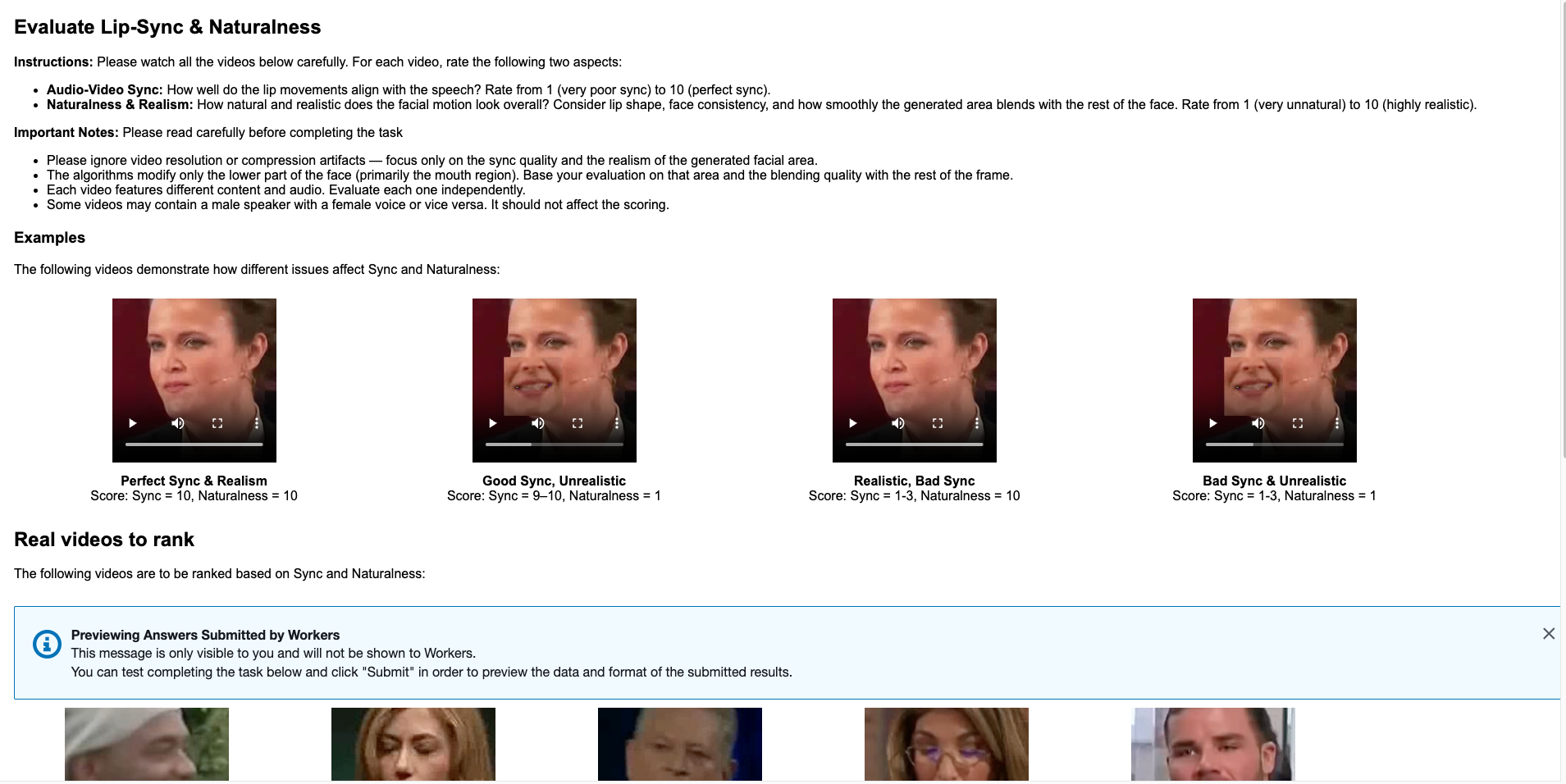}
\caption{Instructions page shown before each evaluation task.}
\end{subfigure}
\vspace{1em}
\begin{subfigure}[b]{0.8\linewidth}
\includegraphics[width=\linewidth]{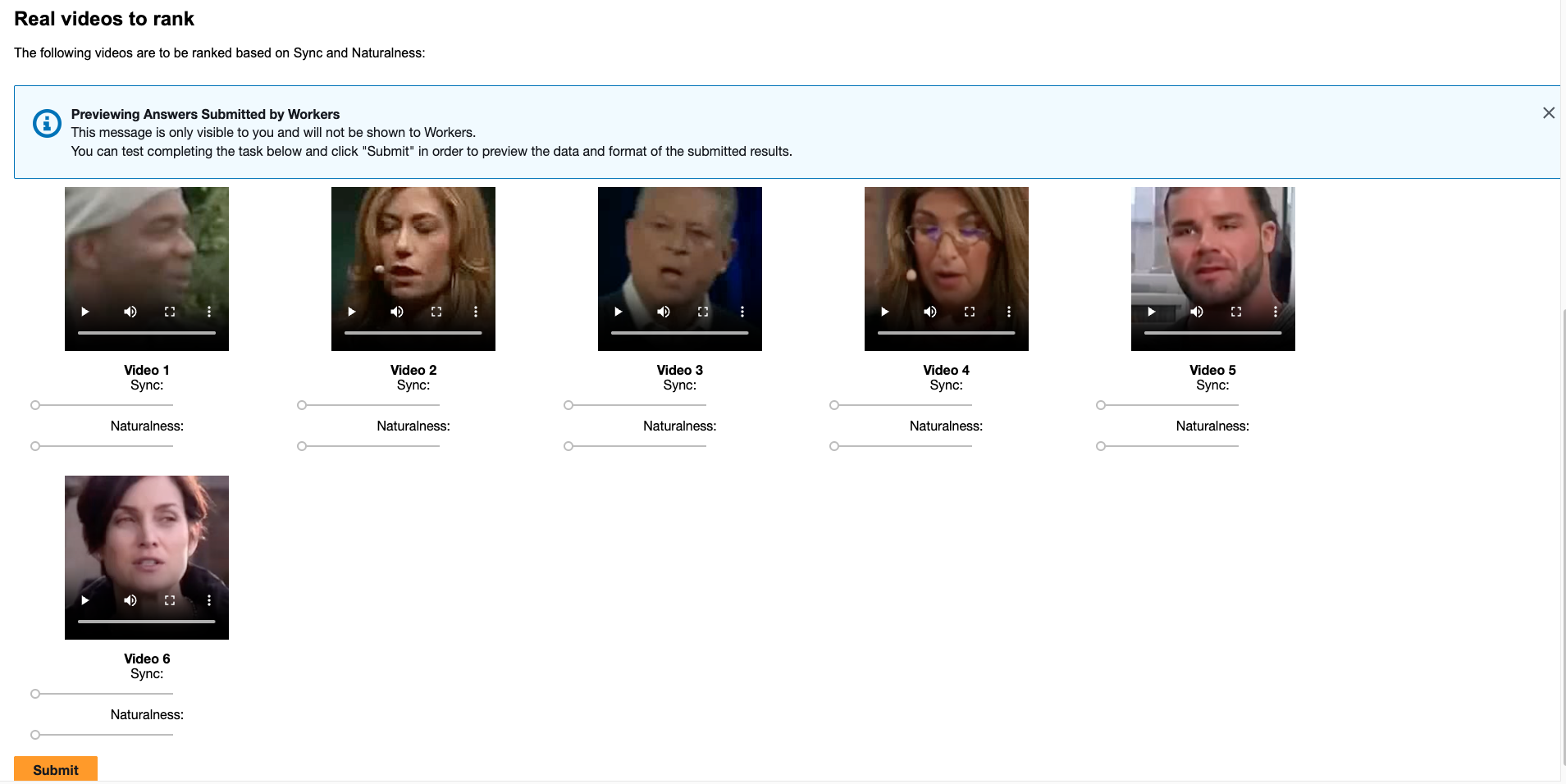}
\caption{Video rating interface showing six videos in randomized order.}
\end{subfigure}
\caption{MOS evaluation interface presented to participants. Each task consists of six randomized videos that participants rated for naturalness and audio-visual synchronization.}
\label{fig:mos_task}
\end{figure}

\begin{figure}[t]
\centering
\includegraphics[width=0.5\linewidth]{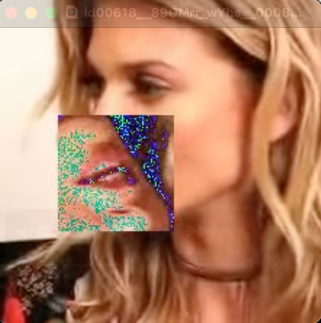}
\caption{Example frame from a corrupted video used for attention screening. The mouth region exhibits visible inconsistencies and noise artifacts.}
\label{fig:corrupted_example}
\end{figure}

To assess the perceptual quality of our method, we conducted a Mean Opinion Score (MOS) test comparing our approach against ground-truth videos and three competing algorithms. Participants evaluated each video on two criteria: (1) \textit{naturalness} and (2) \textit{audio-visual synchronization}.
Each human intelligence task (HIT) presented participants with six videos in randomized order: one ground-truth (GT) video, one synthetically \textit{corrupted} video (used for quality control), and one video from each of the four algorithms (our proposed method and three baselines). The randomization of video presentation mitigated potential ordering bias. Figure~\ref{fig:mos_task} illustrates the evaluation interface: subfigure (a) shows the instruction page presented prior to each task, and subfigure (b) displays the layout for video scoring.
For quality control, we generated corrupted videos by introducing severe visual distortions to the mouth region through the following procedure:
\begin{enumerate}
\item Replacing the mouth region with a constant crop from a different frame, updating this crop once every 25 frames. We used a constant crop to ensure the corrupted video would not inadvertently synchronize with the audio.
\item Rotating the inserted crop by 180 degrees.
\item Adding unique Gaussian noise (mean 0, standard deviation 20) to the pasted region in every frame.
\end{enumerate}
These distortions were deliberately applied only after the first 15 frames and terminated 5 frames before the end, ensuring that participants needed to watch a significant portion of the video to identify the corruption. Figure~\ref{fig:corrupted_example} illustrates an example frame from a corrupted video.
We administered the evaluation using Amazon Mechanical Turk with a compensation of \$0.05 per completed HIT. We collected a total of 4,584 responses, with each response containing scores for six different videos. To ensure data quality, we implemented a screening process that excluded workers who gave average naturalness/synchronization scores lower than 6 for ground-truth videos or higher than 4 for corrupted videos. After applying these quality control measures, we retained 664 valid responses for our final analysis.

\section{Model Architecture and Training Configuration}
This section provides comprehensive details of our model architectures and training procedures to ensure the reproducibility of our results. The full code for training and inference will be added as soon as possible to our anonymous repository: \url{https://github.com/edidub/edidub-results}.

\subsection{Model Components}
Our framework consists of two models, LSD and SRD, each with a 3D-Unet and a diffusion process.
The Lip-Sync Diffusion (LSD) model additionally incorporates an audio quantizer. The implementation of the UNet and diffusion process is based on the official implementation of \citep{dhariwal2021diffusion} as released in \citep{guideddiff}.
\paragraph{Audio Quantizer.}
To generate audio conditioning tokens, we utilize the HuggingFace implementation of HuBERT-Large pretrained on LibriLight60K (model identifier: ``facebook/hubert-large-ll60k''). We discretize HuBERT's output features through k-means clustering applied to 1M feature vectors (extracted from approximately 20,000 seconds of audio) from the single-speaker LJSpeech dataset \citep{ljspeech17}. We configured the clustering with 200 distinct clusters. During inference, each representation vector is mapped to its nearest cluster index, which serves as input to the model.
\paragraph{3D-UNet Architecture.}
Our UNet architecture follows established design principles with targeted modifications as detailed in Table~\ref{tab:unet_hyperparams}.

\FloatBarrier
\begin{table}[H]
\centering
\caption{UNet Architecture Hyperparameters for LSD and SRD. When values differ, they are listed as \textit{LSD | SRD}.}
\label{tab:unet_hyperparams}
\begin{tabular}{ll}
\toprule
\textbf{Hyperparameter} & \textbf{Value} \\
\midrule
Input resolution         & $25 \times 64 \times 64 \times 6$ | $5 \times 224 \times 224 \times 6$ \\
Number of downsampling stages & 4 | 6\\
Base channel dimension   & 64 \\
Channel multipliers      & [1, 2, 4, 8] | [1, 1, 2, 2, 4, 4] \\
Number of residual blocks & 1 per stage \\
Normalization            & GroupNorm (32 groups) \\
Non-linearity            & SiLU \\
Attention layers         & Added at resolutions: [16, 8] | None\\
Attention heads          & 4 \\
Time embedding dimension & 256 \\
Output channels          & 3 (RGB) \\
Skip connections         & Yes (concatenation-based) \\
Conv layers              & True 3D \\
\bottomrule
\end{tabular}
\end{table}

\paragraph{Diffusion Parameters.}
Table \ref{tab:diffusion_hyperparams} presents the hyperparameters for our diffusion processes during both training and inference phases.

\FloatBarrier
\begin{table}[H]
\centering
\caption{Diffusion Process Hyperparameters Used During Training and Inference. When values differ, they are listed as \textit{training | inference}.}
\label{tab:diffusion_hyperparams}
\begin{tabular}{ll}
\toprule
\textbf{Hyperparameter} & \textbf{Value} \\
\midrule
Diffusion type                   & DDPM | DDIM \\
Number of steps                  & 1000 | 50 \\
Scheduler                        & Cosine \cite{nichol2021improved} \\
Timestep spacing                 & Uniform \\
Guidance scale                   & 5.0 \\
Classifier-free guidance         & Yes \\
Stochasticity                    & Fully stochastic | Deterministic (DDIM) \\
Prediction type                  & $\epsilon$ (noise prediction) \\
Multidiffusion \citep{bar2023multidiffusion} window-size       & 24 \\
Multidiffusion step-size         & 12 \\
Clip denoised                    & Yes \\
Noising scheme                   & Inpainting \\
Inversion method                 & DDIM-Inversion \citep{song2020denoising}\\
\bottomrule
\end{tabular}
\end{table}

\subsection{Training Configuration}
Table \ref{tab:training_hyperparams} summarizes the comprehensive set of hyperparameters used during the training of our models.
\FloatBarrier
\begin{table}[H]
\centering
\caption{Training Hyperparameters for LSD and SRD. When values differ, they are listed as \textit{LSD | SRD}}
\label{tab:training_hyperparams}
\begin{tabular}{ll}
\toprule
\textbf{Hyperparameter} & \textbf{Value} \\
\midrule
Optimizer                & AdamW \\
Learning rate            & $1 \times 10^{-4}$ \\
Learning rate scheduler  & Linear warmup + cosine decay \\
Warmup steps             & 10000 \\
Min Learning rate        & $1 \times 10^{-5}$ \\
Batch size               & 28 \\
Gradient accumulation    & 1 \\
Training steps           & 500K \\
EMA (Exponential Moving Average) & Yes, decay $= 0.999$ \\
Weight decay             & 0.01 \\
Loss function            & Masked MSE (MMSE) | MSE \\
Data augmentation        & Horizontal flip | Horizontal flip, Conditioning resolution \\
Masking strategy         & Mask lower-face area | None \\
Mixed precision          & No \\
\bottomrule
\end{tabular}
\end{table}

\section{Benchmarks Curation}
In this section, we provide the video names for each benchmark set. The full pipelines for finding front-faced / occluded videos from VoxCeleb2 can be found in the provided code base. 

\subsection{LRS2 / LRS3}
We randomly sampled 50 utterances from each of the LRS2 and LRS3 corpora, extracting the first 5 seconds of each clip for evaluation. Since LRS3 is partitioned into predefined train, validation, and test splits, our selection is drawn exclusively from its official test set; for LRS2, which does not provide fixed splits, we sample across the entire dataset. The exact video identifiers for both benchmarks are listed in our anonymous repository: \url{https://github.com/edidub/edidub-results}.

\subsection{VoxCeleb2 – Front (Low‐Angle Face Video Selection)}
We developed a procedure to curate videos exhibiting minimal head rotation from the VoxCeleb2 dataset \citep{chung2018voxceleb2}. Our approach leverages MediaPipe’s 3D Face Landmarker \citep{mediapipe} to extract from each sampled frame a $4\times4$ head‐to‐camera transformation matrix, which is then converted into Euler angles (pitch, yaw, roll). As shown in \eqref{eq:front_pose_score}, the pose score ($\alpha_v$) of a video is defined as the maximum absolute angle across all three axes, across all processed frames. Clips shorter than 6 seconds, or those in which no face is detected in a sampled frame, are discarded immediately. Any video whose maximum pose score exceeds $20^\circ$ is rejected; from the remaining pool, we randomly select 50 videos for our benchmark. The complete list of selected filenames is available in our anonymous repository: \url{https://github.com/edidub/edidub-results}.

\begin{equation}\label{eq:front_pose_score}
\alpha_v \;=\;\max_{f \in S(v)} \max\bigl(\,|\phi_f|,\;|\theta_f|,\;|\psi_f|\,\bigr)\,.
\end{equation}

Processing is parallelized across CPU workers to handle the million‐scale utterance set efficiently. Algorithm~\ref{alg:low_angle_selection} details our selection logic, including early exit upon threshold violation and final sorting by ascending pose score.

\begin{algorithm}
\caption{Low‐Angle Face Video Selection}
\label{alg:low_angle_selection}
\begin{algorithmic}[1]
\Require Dataset $V$, min-duration $d_{\min}=6\,$s, angle limit $\theta_{\max}=20^\circ$, frame step $k=5$
\Ensure Selected set $P$
\State $P\gets\emptyset$
\ForAll{$v\in V$}
  \If{Duration$(v)<d_{\min}$} \textbf{continue} \EndIf
  \State $m\gets 0,\;\text{valid}\gets\mathrm{true}$
  \ForAll{frames $f$ at step $k$ in $v$}
    \State $M\gets\text{Detect3DFace}(f)$
    \If{no face in $M$} \State $\text{valid}\gets\mathrm{false}$; \textbf{break} \EndIf
    \State $(\phi,\theta,\psi)\gets\text{EulerAngles}(M)$
    \State $m\gets\max\{m,|\phi|,|\theta|,|\psi|\}$
    \If{$m>\theta_{\max}$} \textbf{break} \EndIf
  \EndFor
  \If{$\text{valid}$ {\bf and} $m\le\theta_{\max}$}
    \State $P\gets P\cup\{(v,m)\}$
  \EndIf
\EndFor
\State \Return $P$ sorted by ascending $m$
\end{algorithmic}
\end{algorithm}

This pipeline enables rapid, large‐scale identification of near‐frontal face clips, ensuring consistent low‐angle pose across our 50‐video benchmark.```

\subsection{VoxCeleb2 - Occluded (Hand-Face Occlusion Video Selection)}

We developed a method to identify videos containing facial occlusions by hands from the VoxCeleb2 dataset \citep{chung2018voxceleb2}. Our approach utilizes MediaPipe's face and hand landmark detection \citep{mediapipe} to quantify occlusion by counting facial landmarks covered by hand regions. After finding occluded videos, we sampled 50 for our benchmark dataset. To recreate our exact benchmark, the videos' names can be found in our anonymous git repo \url{https://github.com/edidub/edidub-results}.

For efficient processing, we sample frames at regular intervals (every 10th frame) and detect both facial landmarks and hand positions. When hands are present, we compute convex hulls around hand landmarks and count the number of facial landmarks falling within these hulls. We track both the total occlusion count and maximum per-frame occlusion across each video. Videos with total occlusion exceeding 30 landmarks are classified as occluded videos.

The processing pipeline is parallelized across available CPU cores, with each worker handling a subset of speakers. Algorithm \ref{alg:occlusion_detection} outlines our video selection procedure:

\begin{algorithm}
\caption{Hand-Face Occlusion Video Selection}
\label{alg:occlusion_detection}
\begin{algorithmic}[1]
\Require Video dataset $V$, occlusion threshold $\theta_{occ}$, frame sampling rate $k$
\Ensure Set of videos $P$ with occlusion metadata
\State $P \gets \emptyset$

\ForAll{video $v \in$ $V$}
    \State $total\_occlusion \gets 0$
    \State $max\_occlusion \gets 0$
    \State $processed\_frames \gets 0$
    
    \ForAll{frame $f$ at every $k$-th position in $v$}
        \State $face\_landmarks \gets$ DetectFaceLandmarks$(f)$
        \If{face detected}
            \State $face\_points \gets$ ConvertToPixelCoordinates$(face\_landmarks)$
            \State $hand\_landmarks \gets$ DetectHandLandmarks$(f)$
            \If{hands detected}
                \State $hand\_hulls \gets$ ComputeConvexHulls$(hand\_landmarks)$
                \State $occlusion \gets$ CountOccludedLandmarks$(face\_points, hand\_hulls)$
                \State $total\_occlusion \gets total\_occlusion + occlusion$
                \State $max\_occlusion \gets \max(max\_occlusion, occlusion)$
                \If{$total\_occlusion \geq 100$}
                    \State \textbf{break} \Comment{Early termination}
                \EndIf
            \EndIf
        \EndIf
        \State $processed\_frames \gets processed\_frames + 1$
    \EndFor
    
    \If{$total\_occlusion > \theta_{occ}$}
        \State $P \gets P \cup \{(v, match\_video, total\_occlusion, max\_occlusion)\}$
    \EndIf
\EndFor

\State \Return $P$
\end{algorithmic}
\end{algorithm}

The convex hull of hand landmarks provides an effective approximation of hand coverage area, and counting landmarks within these hulls offers a quantitative measure of occlusion. This approach allows us to efficiently identify natural hand-to-face occlusions in the video dataset while pairing occluded videos with non-occluded counterparts from the same speakers for controlled experimental analysis.

\end{document}